\documentclass[10pt,twocolumn,letterpaper]{article}

\usepackage[pagenumbers]{cvpr} 

\usepackage{xspace}

\newcommand{\TODO}[1]{\textbf{\color{red}[TODO: #1]}}
\newcommand{\modelName}{DiCoDe\xspace}

\newcommand\blfootnote[1]{%
  \begingroup
  \renewcommand\thefootnote{}\footnote{#1}%
  \addtocounter{footnote}{-1}%
  \endgroup
}

\definecolor{cvprblue}{rgb}{0.21,0.49,0.74}
\usepackage[pagebackref,breaklinks,colorlinks,allcolors=cvprblue]{hyperref}

\usepackage{marvosym}

\def\paperID{7739} 
\def\confName{CVPR}
\def\confYear{2025}

\title{\modelName: Diffusion-Compressed Deep Tokens for\\
Autoregressive Video Generation with Language Models}

\author{
Yizhuo Li\textsuperscript{1,2}, Yuying Ge\textsuperscript{2,\Letter}, Yixiao Ge\textsuperscript{2}, Ying Shan\textsuperscript{2}, Ping Luo\textsuperscript{1,\Letter} \\
\textsuperscript{1}The University of Hong Kong, \textsuperscript{2}ARC Lab, Tencent PCG \\
Project Page: {\small \url{liyz15.github.io/DiCoDe}}
}

\usepackage{layouts}
\usepackage{algorithm}
\usepackage{algpseudocode}
\begin{document}
\maketitle
\blfootnote{This paper is partially supported by the National Key R\&D Program of China No.2022ZD0161000.}

\begin{abstract}

Videos are inherently temporal sequences by their very nature. In this work, we explore the potential of modeling videos in a chronological and scalable manner with autoregressive (AR) language models, inspired by their success in natural language processing. We introduce \textbf{DiCoDe}, a novel approach that leverages \textbf{Di}ffusion-\textbf{Co}mpressed \textbf{De}ep Tokens to generate videos with a language model in an autoregressive manner.
Unlike existing methods that employ low-level representations with limited compression rates, DiCoDe utilizes deep tokens with a considerable compression rate (a 1000× reduction in token count). This significant compression is made possible by a tokenizer trained through leveraging the prior knowledge of video diffusion models.
Deep tokens enable DiCoDe to employ vanilla AR language models for video generation, akin to translating one visual ``language'' into another. 
%
By treating videos as temporal sequences, DiCoDe fully harnesses the capabilities of language models for autoregressive generation. DiCoDe is scalable using readily available AR architectures, and is capable of generating videos ranging from a few seconds to one minute using only 4 A100 GPUs for training. 
%
We evaluate DiCoDe both quantitatively and qualitatively, demonstrating that it performs comparably to existing methods in terms of quality while ensuring efficient training. To showcase its scalability, we release a series of DiCoDe configurations with varying parameter sizes and observe a consistent improvement in performance as the model size increases from 100M to 3B.
%
We believe that DiCoDe's exploration in academia represents a promising initial step toward scalable video modeling with AR language models, paving the way for the development of larger and more powerful video generation models.

\end{abstract}    
\section{Introduction}
\label{sec:intro}

Autoregressive (AR) language models based on transformer architectures~\cite{Brown2020LanguageMA,touvron2023llama,le2023bloom,chowdhery2023palm} such as GPT, with predicting the next token as the objective, have dominated generation tasks in natural language processing (NLP) while showcasing remarkable scalability~\cite{kaplan2020scaling}. The unidirectional design of AR models aligns naturally with the sequential nature of language, where each token depends solely on its predecessors. Unlike other unidirectional models such as RNNs~\cite{grossberg2013recurrent,hochreiter1997long}, transformers exhibit greater scalability due to their parallel trainability and capability to handle longer contexts without strict Markovian constraints. This combination of unidirectional design and exceptional scalability makes AR models the preferred choice for generative tasks in NLP, where text is structured as a sequence of interconnected tokens.

    \begin{figure}
        \centering
        \includegraphics[width=0.9\linewidth]{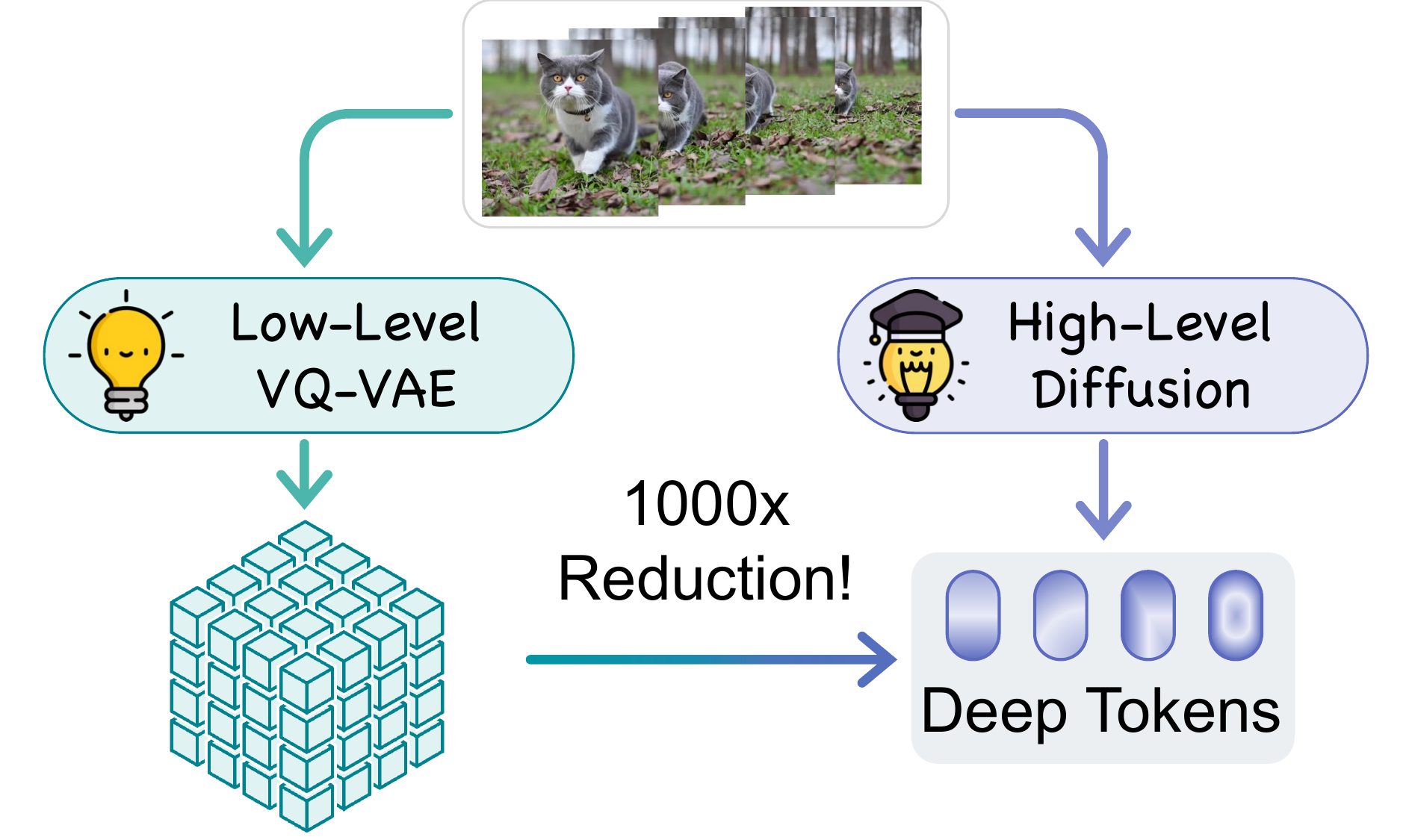}
        \caption{A VQ-VAE-based encoder~\cite{yu2021vector} represent a 2-second $256\times 256$ video clip consisting of 16 frames with 16,384 low-level tokens. Harvesting the prior knowledge of video diffusion models, \modelName compresses the same video clip into 16 high-level deep tokens, achieving \textbf{a 1000$\times$ reduction} in token count. This extremely high compression rate facilitates \textit{chronological video modeling with autoregressive language models}.}
        \label{fig:teaser}
        \vspace{-0.5cm}
    \end{figure}
    
\textit{Similarly, videos can be viewed as sequential processes, akin to language}. However, prevailing methods~\cite{blattmann2023stable,chen2023videocrafter1,chen2024videocrafter2,guo2023animatediff,xing2025dynamicrafter, yang2024cogvideox} for video generation often do not exploit this temporal characteristic. Instead, they tend to treat videos as fixed-length clips generated simultaneously within a diffusion model~\cite{rombach2021highresolution}. While these approaches achieve satisfactory results for generating short clips, they do not accurately reflect the true nature of videos, leading to limitations in scalability, particularly when extending the time dimension. In scenarios involving abrupt scene transitions, these existing methods are likely to falter, underscoring the need for a more sophisticated approach that embraces the inherent temporal structure of videos for generation.

Consequently, the following question naturally arises: \textit{Can videos be modeled in a chronological and scalable manner with autoregressive language models, replicating their success in NLP?} While the analogy between videos and language may seem straightforward at first glance, the redundancy inherent in video data poses a significant challenge. Recent work~\cite{wang2024emu3} has explored the use of language models for autoregressive video generation by tokenizing video clips into low-level discrete tokens produced by a VQ-VAE~\cite{yu2021vector}-based architecture, where a four-frame clip is represented by 4096 discrete codes. With this design, generating a minute-long video may necessitate a context window of up to a million tokens, which is obviously both infeasible and unaffordable. Therefore, there is a pressing need for a tokenizer that can condense video data into high-level tokens with a substantial compression rate.

In this work, we introduce \textbf{\modelName}, a novel approach that leverages \textbf{Di}ffusion-\textbf{Co}mpressed \textbf{De}ep Tokens to generate videos autoregressively with a language model in a chronological and scalable manner. By leveraging the prior knowledge of the video diffusion model~\cite{xing2025dynamicrafter}, DiCoDe learns a frame-level tokenizer to encode fixed-length video clips into high-level tokens, trained through clip-level denoising. In DiCoDe, a 2-second video clip is represented by 16 continuous tokens, which achieves a compression rate that is 1,000$\times$ higher than that of low-level discrete tokens. This significant reduction in token count makes it feasible to model videos temporally with AR language models. These learned deep tokens essentially serve as a ``language'' for videos and are designed to satisfy the following properties: 
1) \textbf{Temporally causal}: By encoding video clips in a way that preserves temporal order, DiCoDe aligns with the sequential nature of AR models and video data;
2) \textbf{Highly compressed}: By leveraging the prior knowledge of video diffusion model, videos can be represented with a manageable number of tokens for efficient AR modeling; 
3) \textbf{Compatible with image data}: Our frame-level tokenizer allows images to be effectively represented, alleviating the scarcity of high-quality video-text data.

With this chronological and compact representation, \modelName employs vanilla AR language models for video generation. To fully unleash the scalability of AR models and utilize well-established architectures, \modelName does not rely on specialized designs like masking strategies with bidirectional attention in previous visual autoregression work~\cite{li2024autoregressive, xie2024show}. Instead, \modelName treats video generation as a straightforward translation task, sequentially generating video tokens based on a text prompt given as the prefix. 
However, the original cross entropy loss in language models for training discrete tokens does not directly apply to continuous token modeling. Recent work~\cite{li2024autoregressive} highlights that AR models are required to model the probability distribution for effective autoregression in a continuous-valued space. This necessity arises from the need to capture the variance of the data rather than relying solely on deterministic modeling. Inspired by this insight, we propose using a Gaussian Mixture Model (GMM) to model the uncertainty of the deep tokens, incorporating variance learning during the autoregressive process. The GMM modeling can be seamlessly integrated as a loss function with minimal modification to existing AR language models, providing a scalable and efficient solution for video generation.

We evaluate \modelName both quantitatively and qualitatively. First, we validate the video tokenization process, demonstrating that \modelName effectively compresses videos into high-level tokens with a significant compression rate and minimal quality degradation. We compare \modelName with established methods on zero-shot video generation tasks in terms of FVD~\cite{unterthiner2018towards} and CLIPSIM~\cite{wu2021godiva}. The experiment results indicate that \modelName achieves performance comparable to existing AR methods while utilizing significantly fewer computational resources (\textit{i.e.}, 4 A100 GPUs). To further demonstrate its scalability, we train a series of \modelName configurations, ranging from 100M to 3B parameters, and observe a consistent performance improvement as the model size increases. 
The effectiveness of \modelName in video generation highlights the vast potential of AR models for temporally sequential video modeling. We hope that our promising initial step will draw increased attention to scalable video modeling using autoregressive language models.

\section{Related Work}
\label{sec:related}

\paragraph{Autoregressive Language Models in Visual Generation}

Autoregressive language models are currently emerging in both image and video generation.
To match with language models, most methods~\cite{wang2024emu3,kondratyuk2023videopoet,villegas2022phenaki,yu2023language,esser2021taming,ramesh2021zero,ding2021cogview,yan2021videogpt,ge2022long,li2024arlon,wang2024loong} tokenize visual data into discrete tokens using methods like VQ-VAE~\cite{yu2021vector}.
Autoregressive video generation models like MAGVIT-v2~\cite{yu2023language} introduces lookup-free quantization to expanding the size of the codebook, achieving exceptional performance in autoregressive video generation.
Recent work~\cite{tschannen2025givt,li2024autoregressive} propose to autoregressively learn continuous tokens, but still relies on low-level VAE latent.
These methods are not applicable to model video in a sequential and scalable manner due to the massive number of tokens required to represent long videos.
\paragraph{Video Diffusion Models}

\begin{figure*}[h]
    \includegraphics{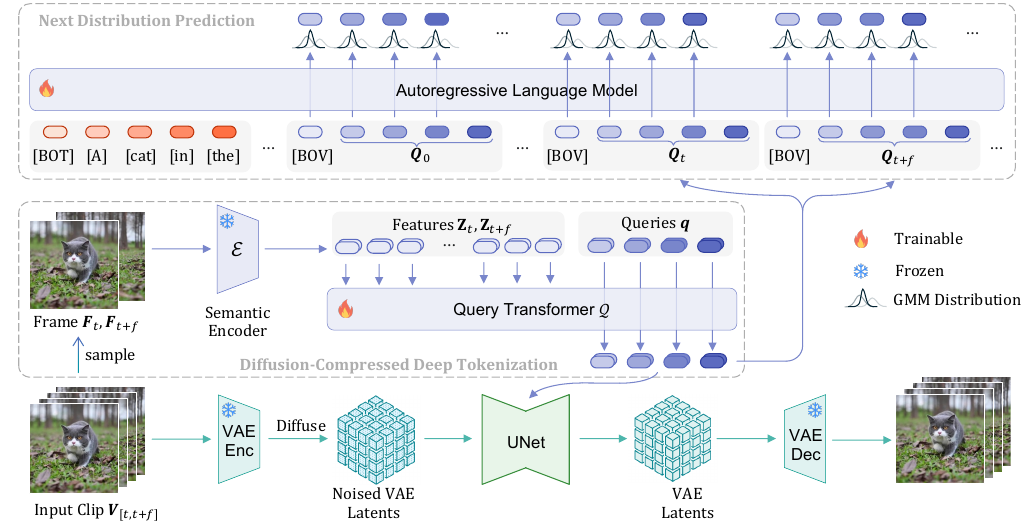}
    \caption{The overall framework of DiCoDe, which consists of a video diffusion model as the tokenizer to extract highly-compressed deep tokens and an autoregressive language model to predict the sequence of deep tokens through modeling distributions.}
    \label{fig:method}
    \vspace{-3mm}
\end{figure*}

Diffusion models are the prevalent method for video generation.
Most diffusion models utilize 3D U-Net or extend the UNet in T2I models with temporal layers.~\cite{ho2022video,ho2022imagen,singer2022make,chen2023videocrafter1,zhou2022magicvideo,wang2023lavie,blattmann2023stable,guo2023animatediff,zeng2024make}
Recently, DiT~\cite{peebles2023scalable}-based diffusion models show promising results in video generation~\cite{ma2024latte,yang2024cogvideox}.
Diffusion models are trained on fixed-length video clips like 16 frames.
Despite diffusion models can be used in autoregressive manner to generate long videos~\cite{henschel2024streamingt2v,qiu2023freenoise,lu2024freelong,kim2024fifo}, they are limited by the receptive field and encounter consistency issues in long video generation.
In our work, we utilized fixed-length diffusion models as a high-ratio compressor with prior world knowledge, and leaves global dynamics and consistency to the autoregressive language model.

\section{Method}
As illustrated in Fig.~\ref{fig:method}, DiCoDe is composed a video diffusion model as the tokenizer to extract deep tokens and a autoregressive language model to predict the sequences of deep tokens.
We first explain the rationale behind the design of deep tokens in Sec.~\ref{sec:language}. In Sec.~\ref{sec:diffusion}, we describe how video diffusion models are utilized to learn the deep tokens. Sec.~\ref{sec:ar} writes about how vanilla autoregressive language models are used to model the sequences of deep tokens. And finally in Sec.~\ref{sec:variance}, variance is introduced via GMMs to facilitate the learning of AR models.

\subsection{Designing a Language for Videos}
\label{sec:language}

This ``language'', i.e. deep tokens, serves as a proxy between the video diffusion model and the AR model.
We justify the design of deep tokens from both cognitive and practical perspectives.
The following principles must be adhered to: 
\textbf{i)} Sequential in time.
To match the sequential nature of AR models and video data, deep tokens must be chronological, which is crucial for naturally extending to longer videos and enhancing scalability.
\textbf{ii)} Highly compressed. 
While an adult reads 200-300 words per minute in English~\cite{brysbaert2019many}, a one-minute $256\times 256$ video of 8FPS typically requires $8 \text{FPS} \times 256 \text{Tokens / Frame} \times 60s \approx 120\text{k}$ tokens to model, which is obviously unmanageable for AR models. 
From a cognitive point of view, modeling videos analogy to language requires an extremely compressed tokenization method.
\textbf{iii)} Compatible with image data.
High-quality video-text data is scarce compared to image-text data, with commonly used datasets like WebVid~\cite{bain2021frozen} are significantly smaller in magnitude than image-text datasets such as LAION~\cite{schuhmann2022laion}. 
However, autoregressive models are data-hungry. 
To mitigate the scarcity of video-text data, deep tokens should integrate seamlessly with image data, allowing the model to leverage the abundant image-text data for much richer semantic information.

Given a video of varying length $\mathbf{V} \in \mathbb{R}^{T\times H\times W\times 3}$, \modelName first sample it with a fixed large frame rate $f$, achieving an $f\times$ reduction in temporal.
The sampled video $\mathbf{V} \in \mathbb{R}^{T/f\times H\times W\times 3}$ is split into frames $\mathbf{F}_0, \mathbf{F}_f, \dots, \mathbf{F}_{T}$. 
The frames are then encoded into high-level semantic-rich tokens individually using an image encoder $\mathcal{E}$ as feature $\mathbf{Z}_t\in \mathbb{R}^{N\times C}$, where $t$ is the frame index, $N$ is the number of tokens, $C$ is the channel dimension. 
Despite $N$ is usually already smaller than the number of typical low-level tokens, it is still too large to fit into AR models if $T$ is large. 
The high-level tokens are further compressed into deep tokens using a query transformer $\mathcal{Q}$. Specifically, $\mathcal{Q}$ learn a fixed set of $N_q$ queries $\mathbf{q} \in \mathbb{R}^{N_q\times C}$ by concatenating $[\mathbf{q}, \mathbf{Z}_t]$ and passing them through a multi-layer self-attention transformer.
The final output of $\mathcal{Q}$ is a set of $N_q$ high-level continuous tokens $\mathbf{Q}_t \in \mathbb{R}^{N_q\times C}$, which is used as the deep tokens.

This design encodes videos in a frame-wise manner, thus to be chronological causal and compatible with image data.
Further sampling in temporal number and compressing in token count using the query transformer achieve an extremely high compression rate.
Encoder $\mathcal{E}$ and query transformer $\mathcal{Q}$ construct a mapping $p(q|v)$ from data space $\mathbf{V} \sim p_{\text{data}}(v)$ to a high-level continuous space $\mathbf{Q} \sim p_{\text{high}}(q)$. \modelName learns this mapping via video diffusion models.

\subsection{Video Diffusion Models as the Tokenizer}
\label{sec:diffusion}

Conventional tokenziers like VAE, VQ-VAE, or VQ-GAN cannot achieve a such high compression rate.
DiCoDe turns to video diffusion models as the tokenizer with rich prior knowledge for an extremely high compression ratio.
The frame-level encoding and clip-level decoding of deep tokens can meet the design principles and achieve satisfying reconstruction quality at the same time.

Given two consecutive sets of deep tokens $\mathbf{Q}_s$ and $\mathbf{Q}_e$ as condition, where $e=s+f$, we train a video diffusion model to denoise Gaussian noises $\epsilon \sim \mathcal{N}(\mathbf{0}, \mathbf{I})$ to the ground-truth video clip $\mathbf{V}_{[s,e]} \in \mathbb{R}^{f \times H\times W\times 3}$. The diffusion loss is defined as
\begin{equation}
    \mathcal{L}(\mathbf{V}_{[s,e]}, \mathbf{Q}_s, \mathbf{Q}_{e})= \mathbb{E}_{\epsilon, t} \left[ \left\| \epsilon - \epsilon_\theta(\mathbf{V}_{[s,e], t}, t, \mathbf{Q}_s, \mathbf{Q}_{e}) \right\|^2 \right]
\end{equation}
where $\epsilon_\theta$ is the prediction of the diffusion model with learnable parameters $\theta$ conditioned on the noised input $\mathbf{V}_{[s,e], t}$, denoising timestep $t$, and deep tokens $\mathbf{Q}_s$ and $\mathbf{Q}_e$.


The mapping $p(v|q)$ essential reconstructs a short video clip from the deep tokens of its head and rear frames.
We make an assumption that video is so redundant that a short clip (e.g., 2 seconds) can be reconstructed solely from its head and rear frames. 
This assumption is required by the temporally downsampling design in Sec.~\ref{sec:language} and can be satisfied with a powerful video diffusion model, leveraging its world knowledge from massive pre-training data.
We validate this assumption in Sec.~\ref{sec:qualitative}. 
\modelName does not apply any other conditions to the video diffusion model such as text or preceding frames \textit{even if it may improve the performance}.
We employ video diffusion model as a high-level tokenizer and leave the generation to AR models completely.

\subsection{Modeling Videos with Autoregressive Models}
\label{sec:ar}

To make the most of the scalability and mature techniques of AR language models, \modelName employs a vanilla autoregressive transformer for video generation.
Given a sequence of sets of deep tokens, we denote $\{\mathbf{T}, Q_{(0, 0)}, \dots, Q_{0, N_q-1}, Q_{f, 0}, \dots,$ $Q_{(t,m)}\}$ as $\mathbf{Q}_{[n,m]}$, where $\mathbf{T}$ is the set of conditional text tokens, $m$ is the token index of the set. We add a $\text{[BOV]}$ token at the beginning of each frame. \modelName autoregressively generates next token according to the temporal order. $\mathbf{Q}_{[n,m]}$ is generated as
\begin{equation}
    \begin{cases}
        p(\mathbf{Q}_{[n,m-1]}) \times  p(Q_{(n,m)} | \mathbf{Q}_{[n,m-1]}) & \text{if } m > 0, \\
        p(\mathbf{Q}_{[n-1, N_q - 1]})  \times p(Q_{(n,0)} | \mathbf{Q}_{[n-1, N_q - 1]}) & \text{if } m = 0.
    \end{cases}
\end{equation}

\modelName does not apply any special design for deep tokens such as masking or bidirectional modeling, which are found not helpful in our settings. 
We attribute this to the compact nature of deep tokens.
Unlike VAE latents or discrete tokens, deep tokens are already highly compressed and semantically rich, thus do not rely heavily on each other for reconstruction.
\modelName treats video generation as a translation task.
This simple choice of AR models allows the utilization of readily available AR architectures and pre-trained models, even if they are designed for text data.

\subsection{Explicitly Introducing the Variance}
Due to the decoupled and offline essences of diffusion loss, the target of the AR model is deterministic.
Unlike in discrete scenario where the target is a categorical distribution, the target in continuous scenario is a point estimate.
As found in MAR~\cite{li2024autoregressive}, directly employing $\mathcal{L}_2$ loss will result in disastrous performance.
An informal justification is that $\mathcal{L}_2$ loss implicitly assumes the target variable follows a Gaussian distribution with a fixed variance.
Specifically, if  the target is Gaussian and $p_{\text{high}}(q)$ $\sim \mathcal{N}(\mu, \sigma^2)$, where $\mu$ is the mean and $\sigma$ is the standard deviation, minimizing the $\mathcal{L}_2$ loss is equivalent to minimizing the negative log-likelihood of the target distributed, which is 
$\mathcal{L}_{\text{NLL}} = \frac{1}{2\sigma^2}||q-\mu||^2 + \text{constant}$.
Since no constraints are applied to the deep tokens, the assumption of a Gaussian distribution does not hold in our case.

We demonstrate that it is possible to explicitly introduce variance to the target even when the target is learned deterministically. 
Instead of directly predicting $\mathbf{Q}$ using the $\mathcal{L}_2$ Loss, \modelName predicts the parameters $\mathcal{P}$ of a variational model, forming a predicted distribution $p_\mathcal{P} (q)$.
During training, the loss is defined as the negative log-likelihood of the target under the predicted distribution
    $ \mathcal{L_{\text{NLL}}} = -\log p_\mathcal{P}(Q) $.
At inference, \modelName samples from the predicted distribution $p_\mathcal{P} (q)$ to generate the next token.

The choice of the pre-defined variational model is flexible. 
We investigate two types of variational models including the Gaussian model and Gaussian Mixture Model (GMM) referred to as Gaussian loss and GMM loss, respectively.
For a simple Gaussian model, \modelName predicts the mean $\mu_{(n, m)}$ and the standard deviation $\sigma_{(n, m)}$ for each $Q_{(n, m)}$.
For a k-component GMM, \modelName predicts $k$ sets of $\mu_{(n, m)}$ and $\sigma_{(n, m)}$ as the mean and standard deviation of each component, and $w_{(n, m)}$ as the weights of each component. 
The reparameterization trick is applied to sample from the predicted distributions in training. One of the components is sampled in inference.
We provide an ablation study on the choice of the target distribution in Section~\ref{sec:ablation}, demonstrating the impact of different variational models on the performance of \modelName.

\label{sec:variance}

\section{Experiments}

\begin{table}[tp]
    \small
    \centering
    \caption{Datasets used for training the AR language models. Only a small portion of image data (a total of 25M) is used.}
    \label{tab:dataset} 
    \begin{tabular}{ccc}
        \toprule
        Name & Type & Num Pairs \\ \midrule
        JourneyDB~\cite{sun2024journeydb} & Image-Text & 4M \\
        Unsplash~\cite{unsplashdata} & Image-Text & 2M \\ 
        Laion-aesthetic-v2-6.25~\cite{schuhmann2022laion} & Image-Text & 1.2B \\  
        LAION-COCO~\cite{schuhmann2022laion} & Image-Text & 600M \\
        WebVid-10M~\cite{bain2021frozen} & Video-Text & 10M \\
        \bottomrule
    \end{tabular}
\end{table}

\subsection{Implementation Details}
\paragraph{Architecture.}
The tokenizer can be implemented with any conditional video generation models.
We use the off-the-shelf video diffusion model DynamiCrafter~\cite{xing2025dynamicrafter}.
DynamiCrafter is a triple-conditioned model conditioned with a text prompt, global visual condition, and full image condition.
To align with the design of \modelName, we remove the text prompt and full image condition, and learns the deep tokens as global visual condition.
We use the pre-trained 256x256 resolution DynamiCrafter, which generates 2s video clips at 8fps at a time.
Except for the ablation study, we use 16 1024-dim tokens for 2s video, i.e. DynamiCrafter is conditioned on merely 32 tokens.
CLIP-ViT~\cite{radford2021learning} are used as the semantic encoder.
Benefiting from the design of deep tokens, we can leverage the existing powerful language models for AR models.
To demonstrate the language knowledge transferability, we use two families of pre-trained AR models: GPT2 and Llama3.2. 
Different sizes of models are used to demonstrate the scalability of \modelName.
For GPT2, we use GPT2(117M), GPT2-Medium(345M), GPT2-Large(762M).
For Llama3.2, we use Llama3.2-1B(1.23B) and Llama3.2-3B(3.3B).
We use the language models' tokenizers to tokenize the text prompt to 80 tokens.
For GMM loss, we use 16 components, and we only need to modify the last projection layer of the AR models to predict $16\times 1024\times 2 + 16=32784$-dim features.
We use \modelName-Llama3.2-1B as the default model unless otherwise specified.

\vspace{-0.5cm}
\paragraph{Datasets.}
For the tokenizer, we use the WebVid-10M dataset with 10M video clips. 
For the AR models, we use a mix of image and video datasets for richer knowledge listed in Tab.~\ref{tab:dataset}.
We use the mix of large-scale of datasets for diversity and only randomly sample \textit{a small portion} of the samples (25M) without repetition since we are already using powerful pre-trained models and limited by the computational resources. We also filter a subset of high motion videos from WebVid-10M using the optical flow.

\vspace{-5mm}
\paragraph{Training.} For the training of video diffusion model, we train on WebVid-10M for 100k iterations with a batch size of 64, using a fixed learning rate of 1e-5, v-prediction and AdamW~\cite{loshchilov2017decoupled} optimizer.
For the training of AR models, we train in a progressive fashion for faster convergence. The model is first trained on image-text datasets for 100k iterations, then trained on the mix of image and video datasets for another 100k iterations, and finally trained on the filtered motion videos for 20k iterations. 
For each video, we sample 256 frames with duration around 32s. We use a global batch size of 256, cosine-scheduled learning rate starting at 1e-4 with warm-up for 1k iterations, and AdamW optimizer.

\begin{figure*}[h]
    \includegraphics[width=\linewidth]{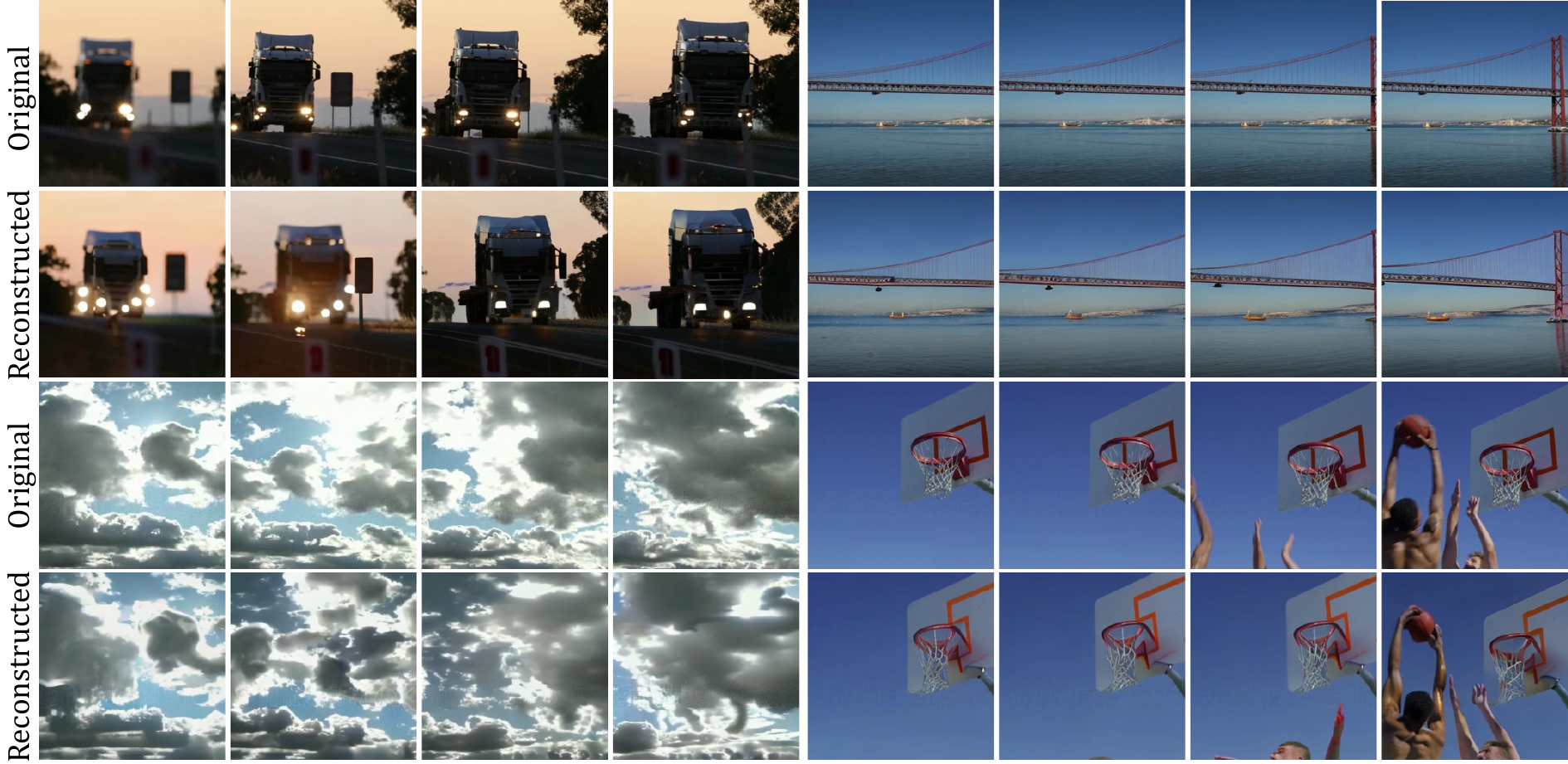}
    \caption{Results of tokenization. A 2-second video clip can be reconstructed effectively from merely 2 frames with 16 deep tokens each, even with object motion (left-top), camera motion (right-top), complex scenes (left-bottom), and emerging entities (right-bottom).}
    \label{fig:tokenization}
\end{figure*}


\begin{table}[tp]
    \small
    \centering
    \caption{Zero-shot video generation results on MSR-VTT dataset.}
    \label{tab:msrvtt}
    \begin{tabular}{lcc}
        \toprule
        Model         &  CLIPSIM $\uparrow$ & FVD $\downarrow$ \\ 
        \midrule
        CogVideo~\cite{hong2022cogvideo}      & 0.2631 &  1294      \\
        ModelScopeT2V~\cite{wang2023modelscope} & 0.2930 &  550       \\ 
        Show-1~\cite{zhang2024show}        & \textbf{0.3072} &  538       \\  
        VideoPoet~\cite{kondratyuk2023videopoet}     & 0.3049 &  \textbf{213}       \\
        Loong~\cite{wang2024loong}        & 0.2903 &  274       \\
        \midrule
        \modelName-Llama3.2-1B    & 0.2950 &  367       \\
        \bottomrule
    \end{tabular}
    \vspace{-1.3mm}
\end{table}

\subsection{Quantitative Results}
To provide a quantitative evaluation of \modelName, we evaluate zero-shot video generation task on MSR-VTT~\cite{xu2016msr} dataset and show the results in Tab.~\ref{tab:msrvtt}. We use the full set of 2990 test videos and randomly sample one caption for each video. We report CLIPSIM~\cite{radford2021learning} and FVD~\cite{unterthiner2019fvd} of Llama3.2-1B as the metrics for comparison.

In this 16-frame clip setting, \modelName only predicts 32 tokens for a 256x256 tokens. 
In contrast to our highly compressed design, other methods utilize orders of magnitude more tokens. 
For example, CogVideo~\cite{hong2022cogvideo} requires 6400 tokens at a spatial resolution of 160x160, Loong requires 1024 tokens at 128x128, and VideoPoet requires 1280 tokens at 128x128.
Other methods also have been pre-trained on massive image-text datasets or use powerful pre-trained text-to-image models.
For example, Show-1~\cite{zhang2024show} utilizes DeepFloyd as initialization and VideoPoet~\cite{kondratyuk2023videopoet} is trained on 1B image-text pairs.
\modelName, on the other hand, is trained on a subset of 25M image-text pairs without repetition. 
And the diffusion tokenizer is only conditioned on deep tokens without text conditioning.

Despite the highly compressed design and limited training, \modelName achieves competitive performance compared to other methods. Both CLIPSIM and FVD are comparable to existing state-of-the-art methods.
Unlike other methods that have sufficient image-text pre-training, DiCoDe is trained on image data at a relatively smaller scale.
One can further improve the performance by pre-training on image-text data at larger scale for better performance.

\subsection{Qualitative Results}
\label{sec:qualitative}

\begin{figure}[tp]
    \includegraphics[width=\linewidth]{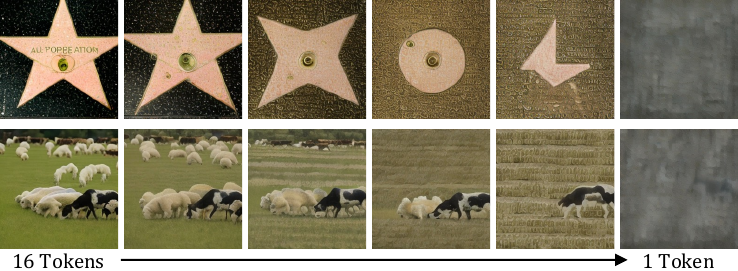}
    \caption{Zering out deep tokens during tokenization reduces the number of entities in reconstructed frames.}
    \vspace{-5mm}
    \label{fig:remove_token}
\end{figure}

\paragraph{Tokenization}
The effectiveness of the diffusion-powered tokenization is shown in Fig.~\ref{fig:tokenization}.
The reconstructed 17-frame 2s videos are generated conditioned on merely 2 frames with 16 deep tokens each.
Despite the extremely high compression ratio, \modelName successfully reconstructs the video clips with minimal degradation.
The results confirm our hypothesis that videos are so redundant that they can be represented with a few deep tokens, even with object motion, camera motion, complex scenes, and emerging entities.
To investigate the essence deep tokens, we gradually zero out deep tokens and show the results in Fig.~\ref{fig:remove_token}.
Removing deep tokens reduces the entities in the video one by one, indicating its rich and condensed semantics, akin to language.

\begin{figure*}[h]
    \centering
    \includegraphics[width=0.95\linewidth]{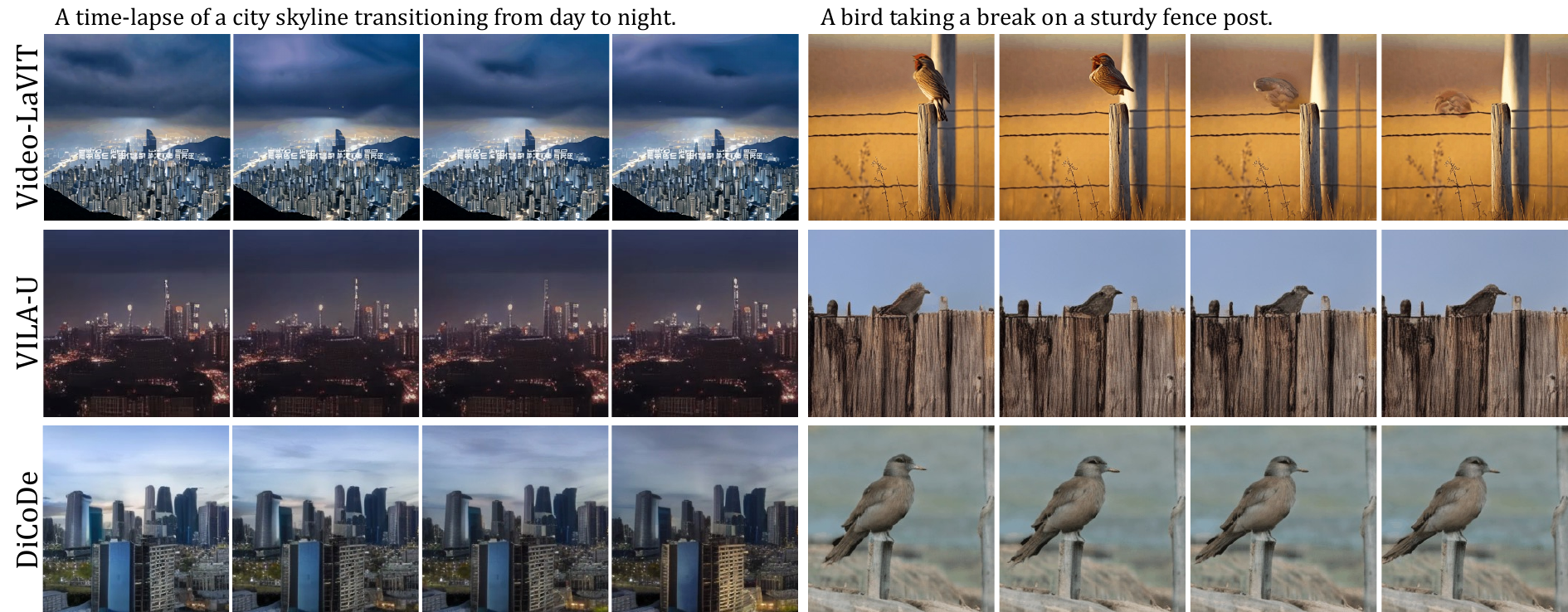}
    \vspace{-0.2cm}
    \caption{Comparison of short video generation results. \modelName generates more coherent, dynamic and prompt-following videos. On the left, it captures the transition from day to night. On the right, it maintains the appearance of the bird across frames with noticeable motion.}
    \label{fig:short_compare}
    \includegraphics[width=0.95\linewidth]{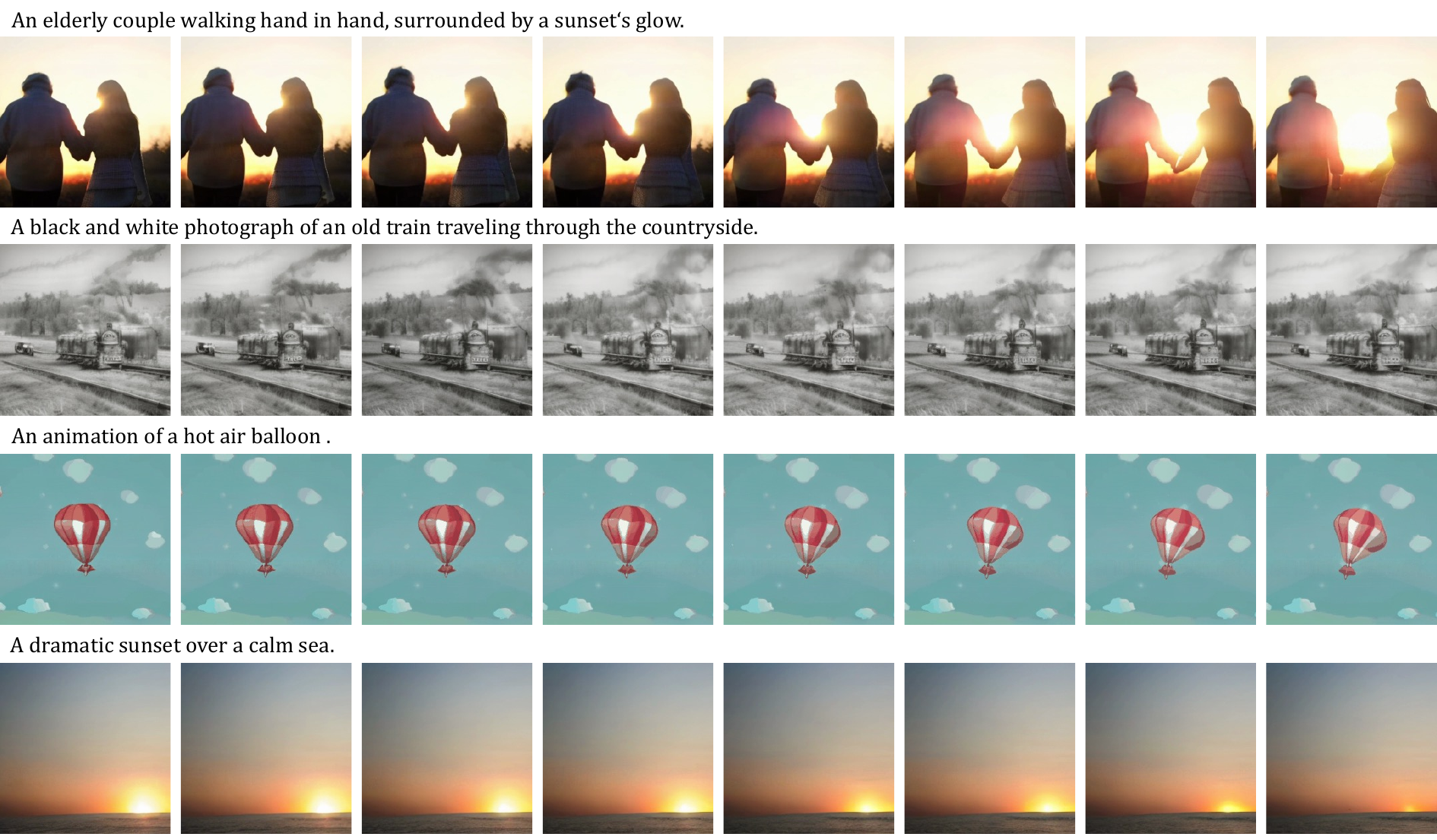}
    \vspace{-0.2cm}
    \caption{Additional results generated by \modelName.}
    \label{fig:short_results}
    \includegraphics[width=0.95\linewidth]{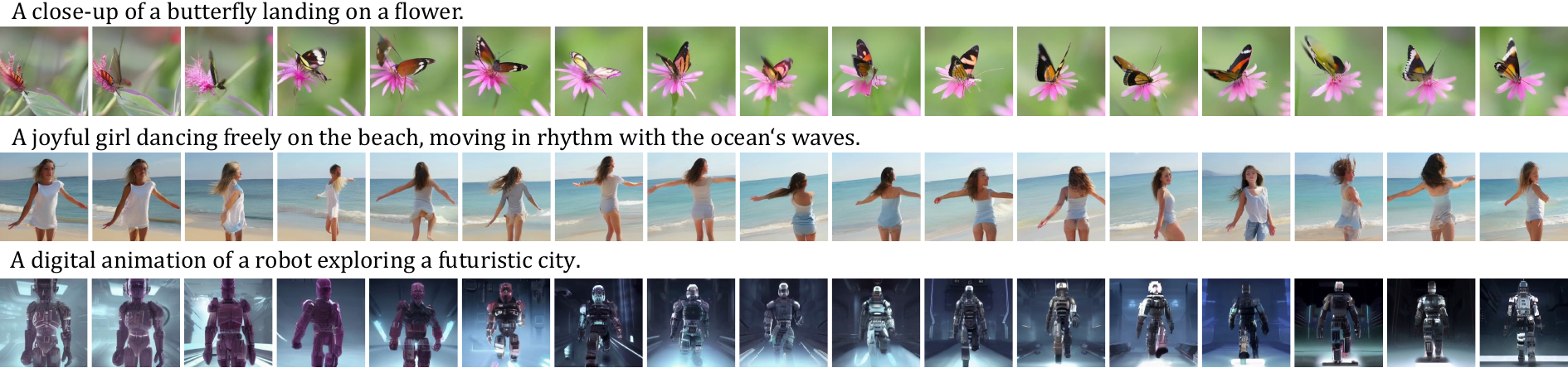}
    \vspace{-0.2cm}
    \caption{Long videos generated by \modelName. Frames are evenly sampled from 256-frames (64s) video. 
    \label{fig:long_results}
    }
\end{figure*}

\vspace{-2mm}
\paragraph{Short Video Generation}
We qualitatively compare the short video generation results of \modelName with other autoregressive methods with examples in Fig.~\ref{fig:short_compare}.
Video-LaVIT~\cite{jin2024video} heavily relies on the pre-trained text-to-image model for visual quality but finds it hard to fully capture motion dynamics. It generates almost static images for time-lapse transition or cannot maintain object appearance.
VILA-U~\cite{wu2024vila} also fails to capture the transition from day to night and generates deformed objects.
Our \modelName generates more coherent, dynamic and prompt-following videos. 
In the examples, it successfully captures the transition from day to night and maintains the appearance of the bird across frames.
More results are illustrated in Fig.~\ref{fig:short_results}.

\begin{figure*}[!ht]
    \centering
    \begin{subfigure}[b]{0.32\textwidth}
        \includegraphics[width=\textwidth]{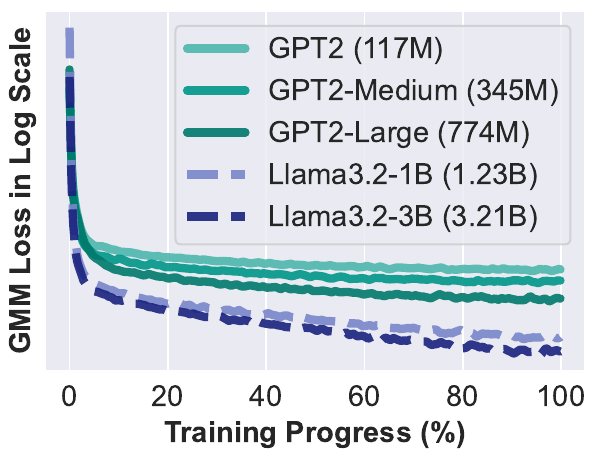}
    \end{subfigure}
    \hfill
    \begin{subfigure}[b]{0.32\textwidth}
        \includegraphics[width=\textwidth]{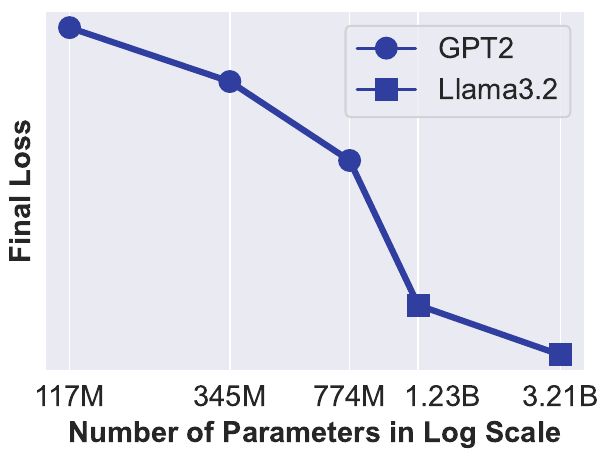}
    \end{subfigure}
    \hfill
    \begin{subfigure}[b]{0.32\textwidth}
        \includegraphics[width=\textwidth]{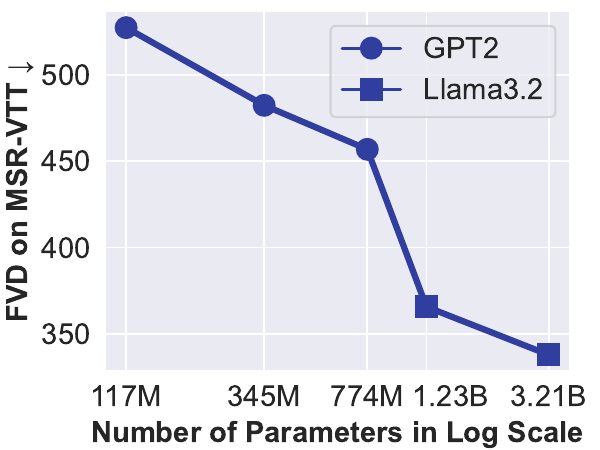}
    \end{subfigure}
    \caption{Ablation on AR model sizes. Larger models achieve lower loss and FVD, demonstrating clear scalability.}
    \label{fig:scalability}
    \vspace{-2mm}
\end{figure*}

\vspace{-2mm}
\paragraph{Long Video Generation}
In Fig.~\ref{fig:long_results}, we show the long video generation results of \modelName.
\modelName can generate long videos seamlessly with consistent appearance and motion dynamics.
Compared to other autoregressive long video generation methods like Loong~\cite{wang2024loong}, \modelName does not require any additional techniques like truncating or prefixing for long video extension, and thus enjoys better dynamics and consistency.
Constrained by data and computational resources, we sample 256 frames at 4 fps (1 minute) for each video.
256 frames only account for 256 visual tokens in the AR model, which is still far shorter than the context length of modern language models, demonstrating the potential for extending longer.

\begin{table}
    \small
    \centering
    \caption{Ablation study on the number of deep tokens, evaluated with zero-shot video generation.}
    \label{tab:token_num}
    \begin{tabular}{lccc}
        \toprule
        Num Deep Tokens & 8   & 16  & 32 \\ 
        \midrule
        FVD on MSR-VTT $\downarrow$ & 600 & \textbf{565} & 641  \\
        \bottomrule
    \end{tabular}
    \vspace{-3mm}
\end{table}

\subsection{Scalability of Autoregressive Language Models}
\label{sec:scalability}

Scalability is the key advantage of employing vanilla AR language models.
We verify the scalability of \modelName across different sizes of AR models in terms of training loss and reconstruction quality.
All experiments use the same setting except for the size of the AR models.
As listed in Fig.~\ref{fig:scalability}, it is obvious that given the same training budget, larger models achieve lower pre-training loss and lower FVD.
There is a clear gap between GPT2 family and Llama3.2 family, which is attributed to the better pre-training of Llama3.2.
This gap confirms that even across modality, the knowledge from text can be transferred to video generation to some extent, showing that video generation is indeed treated as a language generation with the deep tokens design.
We also notice that from the loss curve, Llama3.2 models are far from convergence, indicating potential for further improvement with more training budget.

\subsection{Ablation Studies}
\label{sec:ablation}

\paragraph{Number of Deep Tokens}

We ablate the effect of deep token number for each frame in tokenization including 8, 16, 32 tokens.
Different tokenizers of different tokens are first trained with the same settings.
Then, the tokenizers are used to generate videos with Llama3.2-1B.
Qualitative results show that more tokens lead to better reconstruction quality.
The zero-shot generation results are shown in Tab.~\ref{tab:token_num}.
The FVD is reported after image-video mixed training for ablation purpose.
The results look surprising that the 16-token tokenizer achieves the best FVD while 32-token performs worse than 8-token setting.
However, this just aligns with our deep token design, that autoregressive models accumulates error with length increasing, and the 32-token setting is more error-prone than 16-token setting.
Therefore, for best performance, one need to strike a balance between the reconstruction quality and the compression ratio.

\vspace{-5mm}
\paragraph{Loss Type}

The introduction of variance is a key design of \modelName.
We ablate different loss types for the AR models including $\mathcal{L}_2$ loss, Gaussian loss, and GMM loss with 16 kernels.
The results are shown in Tab.~\ref{tab:loss_type}.
The GMM loss performs the best while the $\mathcal{L}_2$ loss performs the worst.
This is expected as the introduction of variance can better capture the uncertainty in the distribution of the deep tokens, which is crucial for the diversity of generation.

\begin{table}
    \small
    \centering
    \caption{Ablation study on the loss type for training the language model, evaluated with zero-shot video generation.}
    \label{tab:loss_type}
    \begin{tabular}{lccc}
        \toprule
        Loss Type & $L_2$   & Gaussian & 16-GMM \\ 
        \midrule
        FVD on MSR-VTT $\downarrow$ & 643 & 593 & \textbf{565}  \\
        \bottomrule
    \end{tabular}
    \vspace{-3mm}
\end{table}

\section{Conclusion}

We propose \modelName, a video generation framework that models videos in a chronological and scalable manner with autoregressive language models.
With the design of deep tokens, diffusion-powered compression, and variance in the target, \modelName generates videos as temporal sequences, aligning with their sequential nature.
The experiments demonstrate the effectiveness and scalability of \modelName.
We hope \modelName reveal a new paradigm for video generation and inspire the development of larger-scale long video generation models in the future.

{\flushleft \bf Limitation.} The primary limitation comes from the data.
\modelName excels with natural scenes and nonrigid objects such as ocean, mountain, and forest but sometimes struggles with rigid objects due to the WebVid-10M dataset's focus on natural scenes. The length of WebVid-10M (17.5s in average) also heavily restricts the potential for long video generation. Another limitation is the design of deep tokens. While our assumption that video redundancy allows for compression into a few tokens holds in most cases, it may fail with videos that have sudden transitions. 


{
    \small
    \bibliographystyle{ieeenat_fullname}
    \bibliography{main}
}

\clearpage
\setcounter{page}{1}
\maketitlesupplementary

\section{Implementation Details}

\paragraph{Video Diffusion Models}
To adopt DynamiCrafter~\cite{xing2025dynamicrafter} as the deep tokenizer, we remove the text prompt and full image condition by setting text prompt to empty and changing the input channels from 8 to 4.
We start with the $256\times 256$ resolution pretrained model, without loading the first convolutional layer in the UNet.
We use the CLIP-ViT-H/14 from LAION~\cite{schuhmann2022laion} as the semantic encoder $\mathcal{E}$.
We reuse the query transformer in DynamiCrafter as the query transformer $\mathcal{Q}$. 
The query transformer has 4 layers with 12 heads and 1024 hidden dimensions.
In the 16-token setting, we reused the first 16 queries in the pre-trained model's query transformer and remove the rest.
The model is pretrained on WebVid-10M~\cite{bain2021frozen} for 100k steps with a total batch size of 64, a learning rate of 1e-5, and AdamW~\cite{loshchilov2017decoupled} optimizer.
The videos are sampled at video length of 16 and max frame stride of 6, with approximately 2s per video.
We use random frame stride which samples frame stride from 1 to 6 for better motion condition.
The videos are resized and center-cropped to $256\times 256$.
We remove the watermark in the WebVid dataset using template matching.
At inference, we use 50 DDIM sampling steps, CFG (classifier-free guidance) scale of 7.5, and guidance rescale of 0.7.

\paragraph{Autoregressive Language Models}
We use the GPT2 and Llama3.2 models from HuggingFace Transformers as the autoregressive language models.
For GPT2, we use GPT2 of size 117M, GPT2-Medium of size 345M, and GPT2-Large of size 762M.
For Llama3.2, we use Llama3.2-1B of size 1.23B and Llama3.2-3B of size 3.3B.
We utilize each model's original text tokenizer to tokenize the text prompt.
We pad the text prompt to 80 tokens or truncate it if it exceeds 80 tokens.
Given a long video, it is temporally downsampled to FPS 0.5, and individual frames are encoded with the pre-trained semantic encoder $\mathcal{E}$ and the query transformer $\mathcal{Q}$.
Two 3-layer MLPs are used to project the visual tokens to the same dimension as the text tokens and the predicted features to the output dimension respectively.
We add the same [BOV] (begin of the visual) token to the beginning of each 16-token feature.
We also add two sets of learnable positional embeddings to the visual tokens, one for the frame index and one for the token index.
5\% of the visual tokens are dropped during training for potential classifier free guidance.
We use 16 components for the GMM loss by modifying the final projection layer in the autoregressive language models.
For each token, it predicts the mean and variance of the Gaussian distribution, with $16\times 1024 \times 2+16=32784$-dim feature.
The algorithm for computing the GMM loss is shown in Alg.~\ref{alg:gmm_loss}.

\begin{algorithm}[tp]
    \caption{Gaussian Mixture Model Loss Computation}
    \label{alg:gmm_loss}
    \begin{algorithmic}[1]
    \Require Means \( \boldsymbol{\mu} \) of size \( K \times d \), Variances \( \boldsymbol{\sigma}^2 \) of size \( K \times d \), Mixture weights \( \mathbf{w} \) of size \( K \), Target data \( \mathbf{X} \) of size \( N \times d \)
    \Ensure Negative log-likelihood loss \( \text{nll} \)
    
    \Function{GMM Loss}{$\boldsymbol{\mu}$, $\boldsymbol{\sigma}^2$, $\mathbf{w}$, $\mathbf{X}$}
        \State \( d \gets \text{Number of dimensions of } \mathbf{X} \)
        \State \( \boldsymbol{\sigma} \gets \sqrt{\boldsymbol{\sigma}^2} \)
        
        \For{\( n = 1 \) to \( N \)}
            \State \( L_n \gets 0 \)
            \For{\( k = 1 \) to \( K \)}
                \State \( p_{nk} \gets \mathcal{N}(\mathbf{X}_n | \boldsymbol{\mu}_k, \text{diag}(\boldsymbol{\sigma}_k^2)) \)
                \State \( L_n \gets L_n + w_k \times p_{nk} \)
            \EndFor
        \EndFor
        
        \State \( \text{nll} \gets -\frac{1}{N} \sum_{n=1}^{N} \log L_n \)
        
        \State \Return \( \text{nll} \)
    \EndFunction
    \end{algorithmic}
\end{algorithm}

The autoregressive models are trained in a progressive manner for faster convergence.
For image pre-training, it is first trained on image data listed in Tab.~\ref{tab:dataset} for 100k iterations with a batch size of 256 using $\mathcal{L}_2$ Loss, then another 100k iterations using GMM Loss. 
For video training, it is first trained on mixed image and video data for 100k iterations with a batch size of 256 and image:video ratio of 1:1.
Then it is trained on filtered motion videos for another 20k iterations.
To filter the motion videos, we compute the optical flow between consecutive frames and 1 fps and average the flow magnitude as the motion score.
We choose videos with motion score between 0.5 and 1.38, length between 556 and 4000 frames, resulting in 1M videos in total.
The length of the video is limited to 256 frames, with a max context length of $80+(1 + 16)*16=352$, in which the [BOV] token is also counted.
The videos are also filtered with a minimum aspect ratio of 0.3333, minimum resolution 200.
To adapt the video diffusion models to the variance introduced by the autoregressive language models, further fine-tuning the video diffusion models for another 20k iterations based on the prediction of the autoregressive language models.
For all training, we use a learning rate of 1e-4, cosine schedule with warm-up for 1k iterations, AdamW optimizer with $\beta_1=0.9$, $\beta_2=0.98$, $\epsilon=1e-6$, weight decay $0.05$.

\begin{figure*}[h]
    \centering
    \includegraphics[width=\linewidth]{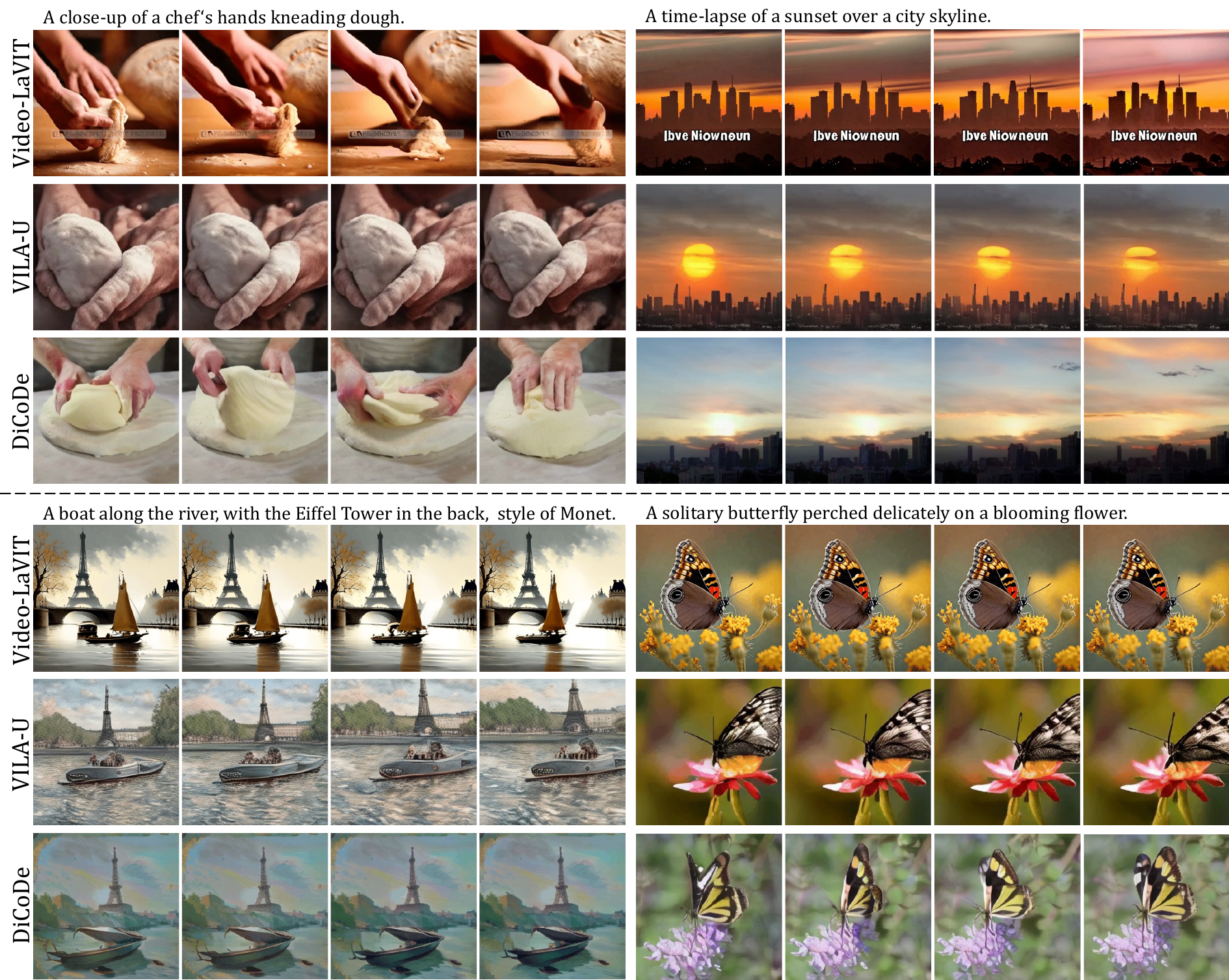}
    \caption{Additional comparison of short video generation results. }
    \label{fig:short_compare_suppl}
\end{figure*}

\section{Additional Qualitative Results}

The videos showcasing qualitative results are available in the attached folder, encompassing both the main content and supplementary materials.

Additional comparisons are illustrated in Fig.~\ref{fig:short_compare_suppl}.
\begin{itemize}
    \item Top-left: Both Video-LaVIT and VILA-U exhibit deformed hand motions, whereas \modelName produces consistent and accurate motion.
    \item Top-right: Video-LaVIT and VILA-U generate nearly static videos, while \modelName effectively captures the transition of the sunset.
    \item Bottom-left: Video-LaVIT’s generation progressively deforms, but \modelName maintains the boat’s consistency across frames.
    \item Bottom-right: Both Video-LaVIT and VILA-U create almost static videos, while \modelName successfully models the butterfly’s dynamic movement.
\end{itemize}

\begin{figure*}[h]
    \centering
    \includegraphics[width=0.95\linewidth]{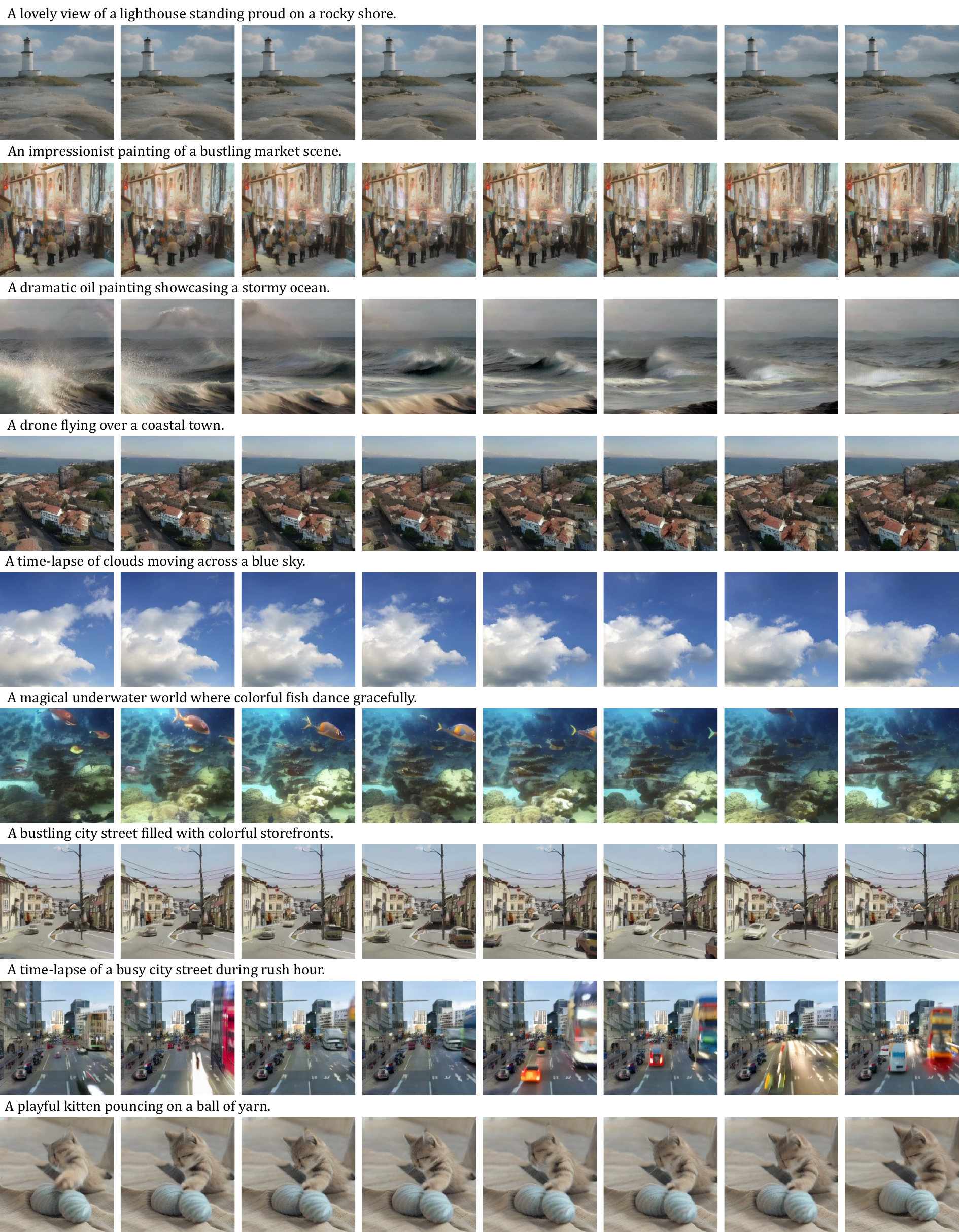}
    \caption{Additional video generation results. }
    \label{fig:short_results_suppl}
\end{figure*}

\begin{figure*}[tp]
    \centering
    \includegraphics[width=\linewidth]{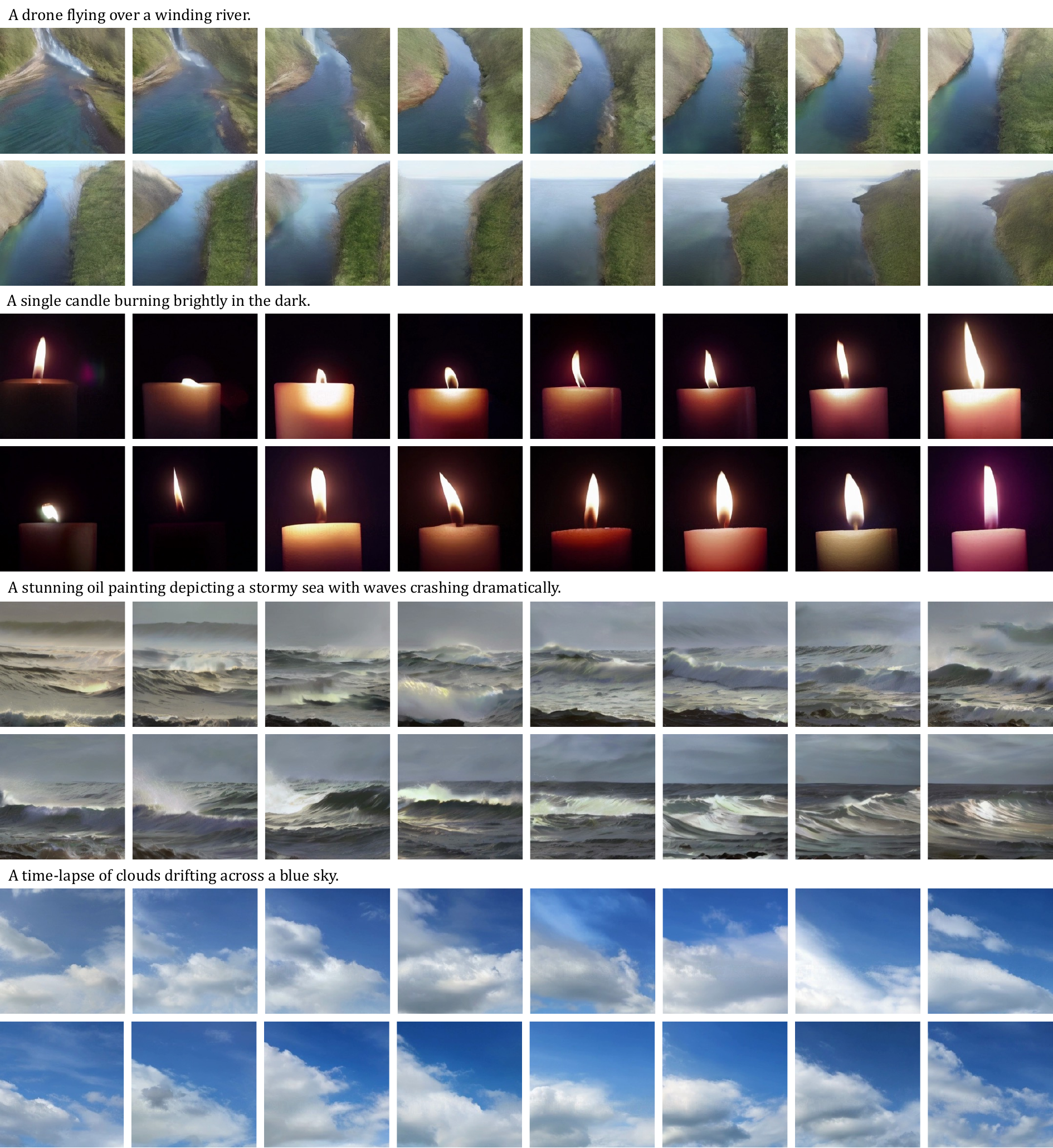}
    \caption{Additional long video generation results. }
    \label{fig:long_results_suppl}
\end{figure*}

Additionally, Fig.\ref{fig:short_results_suppl} and Fig.\ref{fig:long_results_suppl} present further results on short and long video generation, respectively. The outputs from \modelName are dynamic, coherent, and faithfully adhere to the provided prompts.

\end{document}